\documentclass{sig-alternate-2013} 
\newfont{\mycrnotice}{ptmr8t at
  7pt} \newfont{\myconfname}{ptmri8t at 7pt} \let\crnotice\mycrnotice%
\let\confname\myconfname%

\permission{Permission to make digital or hard copies of all or part
  of this work for personal or classroom use is granted without fee
  provided that copies are not made or distributed for profit or
  commercial advantage and that copies bear this notice and the full
  citation on the first page. Copyrights for components of this work
  owned by others than ACM must be honored. Abstracting with credit is
  permitted. To copy otherwise, or republish, to post on servers or to
  redistribute to lists, requires prior specific permission and/or a
  fee. Request permissions from Permissions@acm.org.}
\conferenceinfo{RecSys'15,}{September 16--20, 2015, Vienna, Austria.}
\copyrightetc{\copyright~2015 ACM. ISBN \the\acmcopyr}
\crdata{978-1-4503-3692-5/15/09\ ...\$15.00.\\
  DOI: http://dx.doi.org/10.1145/2792838.2800192}

\usepackage{hyperref} \usepackage{graphicx} \usepackage{multirow}
\usepackage{amsmath} \usepackage{amssymb} \usepackage{color}
\usepackage{booktabs}
\usepackage[utf8]{inputenc}

\newcommand{\bmx}[0]{\begin{bmatrix}}
  \newcommand{\emx}[0]{\end{bmatrix}}

\newcommand{\vect}[1]{\mathbf{#1}}
\newcommand{\vects}[1]{\boldsymbol{#1}}
\newcommand{\matr}[1]{\mathbf{#1}}

\newcommand{\vb}[0]{\vect{b}} 
\newcommand{\vc}[0]{\vect{c}} \newcommand{\vo}[0]{\vect{o}}
 \newcommand{\vh}[0]{\vect{h}}
\newcommand{\vi}[0]{\vect{i}}

\newcommand{\vf}[0]{\vect{f}} \newcommand{\vy}[0]{\vect{y}}

\newcommand{\mE}[0]{\matr{E}} \newcommand{\mW}[0]{\matr{W}}
 
\newcommand{\mQ}[0]{\matr{Q}} \newcommand{\mU}[0]{\matr{U}}
\newcommand{\mV}[0]{\matr{V}} \newcommand{\mA}{\matr{A}}

\newcommand{\TT}[0]{\vects{\theta}}
\newcommand{\vtau}[0]{\vects{\tau}}

\newcommand{\vgamma}[0]{\vects{\gamma}}

\DeclareMathOperator*{\argmin}{\arg \min}

\begin{document}
%

\title{Learning Distributed Representations from Reviews for
  Collaborative Filtering}
%
%
%
%
%

\numberofauthors{1} 
%
\author{
%
%
  \alignauthor
  Amjad Almahairi, Kyle Kastner, Kyunghyun Cho, Aaron Courville\\
  \affaddr{D\'{e}partement d'Informatique et de Recherche Op\'{e}rationelle}\\
  \affaddr{Universit\'{e} de Montr\'{e}al}\\
  \email{\{amjad.almahairi, kyle.kastner, kyunghyun.cho,
    aaron.courville\}@umontreal.ca} }

\maketitle
\begin{abstract}

  Recent work has shown that collaborative filter-based recommender
  systems can be improved by incorporating side information, such as
  natural language reviews, as a way of regularizing the derived
  product representations.  Motivated by the success of this approach,
  we introduce two different models of reviews and study their effect
  on collaborative filtering performance. While the previous
  state-of-the-art approach is based on a latent Dirichlet allocation
  (LDA) model of reviews, the models we explore are neural network
  based: a bag-of-words product-of-experts model and a recurrent
  neural network.  We demonstrate that the increased flexibility
  offered by the product-of-experts model allowed it to achieve
  state-of-the-art performance on the Amazon review dataset,
  outperforming the LDA-based approach. However, interestingly, the
  greater modeling power offered by the recurrent neural network
  appears to undermine the model's ability to act as a regularizer of
  the product representations.
\end{abstract}


\section{Introduction}

Recommendation systems are a crucial component of many e-commerce
enterprises, providing businesses with metrics to direct consumers to
items they may find appealing. A general goal of these systems is to
predict a user's preference for a certain product, often represented
as an integer-valued rating, e.g., between $1$ (unsatisfied) and $5$
(satisfied).

In order to predict the user's preference for a product, it is often
beneficial to consider as many sources of information as possible,
including the preference of the user for other products, the
preferences of other users, as well as any side information such as
characteristics of each user and product. A data-driven approach based
on this idea is called {\it collaborative filtering}.

Collaborative filtering has been successfully used for recommendation
systems (see, e.g., \cite{ricci2011:handbook}). A typical approach to
using collaborative filtering for recommendation systems is to
consider all the observed ratings given by a set of users to a set of
products as elements in a matrix, where the row and column of this
matrix correspond to users and products, respectively. As the observed
ratings is typically only a small subset of the possible ratings (all
users rating all products), this matrix is sparse. The goal of
collaborative filtering is to fill in the missing values of this
matrix: to predict, for each user, the rating of products the user has
not rated. In this setting, collaborative filtering is usually cast as
a problem of matrix factorization with missing
values~\cite{Ilin2010,mnih2007:pmf,salakhutdinov2008:bpmf}. The sparse
matrix is factorized into a product of two matrices of lower rank
representing a user matrix and a product matrix.  Once these matrices
are estimated, a missing observation can be trivially reconstructed by
taking a dot product of a corresponding user vector (or
representation) and a product vector (or representation).

In this formulation of collaborative filtering, an important issue of
data sparsity arises. For instance, the dataset provided as a part of
the Netflix Challenge\footnote{\url{http://www.netflixprize.com/}} had
only 100,480,507 observed ratings out of more than 8 billion possible
ratings\footnote{ 480,189 users and 17,770 movies } (user / product
pairs) meaning that 99\% of the values were missing.  This data
sparsity easily leads to naive matrix factorization overfitting the
training set of observed ratings~\cite{Ilin2010}.



In this paper, we are interested in regularizing the collaborative
filtering matrix factorization using an additional source of
information: reviews written by users in natural language. Recent work
has shown that better rating prediction can be obtained by
incorporating this kind of text-based side
information~\cite{McAuley13:hft,Ling2014:RMR,bao2014topicmf}.
Motivated by these recent successes, here we explore alternative
approaches to exploiting this side information. Specifically, we study
how different models of reviews can impact the performance of the
regularization.

We introduce two approaches to modeling reviews and compare these to
the current state-of-the-art LDA-based approaches
\cite{McAuley13:hft,Ling2014:RMR}. Both models have previously been
studied as neural-network-based document models.  One is based on the
Bag-of-Words Paragraph Vector \cite{Le14:parvec}. This model is
similar to the existing LDA-based model, but, as we argue, it offers a
more flexible natural language model. The other is a recurrent neural
network (RNN) based approach. RNNs have recently become very popular
models of natural language for a wide array of tasks
\cite{Le14:parvec}. Here we will find that despite the considerable
additional modelling power brought by the RNN, it does not offer
better performance when used as a regularizer in this context.

The proposed approaches are empirically evaluated on the Amazon
Reviews Dataset~\cite{McAuley13:hft}. We observe that the proposed
bag-of-words language model outperforms the existing approach based on
latent Dirichlet allocation (LDA,~\cite{blei2003latent}). We also
confirm the use of an RNN language model does not lead to improved
performance.  Overall, our experiments demonstrate that, in this
particular application where we rely on the document model to
regularize the collaborative filtering matrix factorization,
controlling the model flexibility is very important.

We also make methodological contributions in studying the effect of
the train / test splits used in experimentation. Previous works on
this subject (e.g. \cite{McAuley13:hft}, \cite{Ling2014:RMR} and
\cite{bao2014topicmf}), do not clearly identify how the data was split
into train and test sets. Here we empirically demonstrate the
importance of doing so. We show that for a given fixed split,
conclusions regarding the relative performance of competing approaches
do generalize to other splits, but comparing absolute performance
across difference splits is highly problematic.

\section{Matrix Factorization for Collaborative Filtering}
\label{sec:mf}

Let us assume that we are given a set
$R = \left\{ r_{u,i} \right\}_{(u,i) \in O_R}$ of observed ratings,
where $r_{u,i} \in \left\{ 1, 2, \cdots, 5\right\}$ is the rating
given by the user $u$ to the product $i$. Collaborative filtering aims
at building a model that is able to predict the rating of an
unobserved user-product pair, i.e., $r_{u,i}$ where
$(u,i) \notin O_R$.

In collaborative filtering based on matrix factorization, we estimate
each user-product rating as
\begin{align}
  \label{eq:matfact}
  r_{u,i}\approx\hat{r}_{u,i} = \mu+\beta_{u}+\beta_{i}+\vgamma_{u}^{\top}\vgamma_{i},
\end{align}
where $\mu$, $\beta_i$ and $\beta_u$ are a global bias, a
user-specific bias for the user $u$ and a product-specific bias for
the product $i$, respectively.  The vectors $\vgamma_u$ and
$\vgamma_i$ are the latent factors of the user $u$ and the product $i$
respectively.

We estimate all the parameters in the r.h.s of Eq.~\eqref{eq:matfact}
by minimizing the mean-squared error between the predicted ratings and
the observed, true ratings:
\begin{align}
  \label{eq:cost_r}
  C_R(\TT)=\frac{_{1}}{\left|O_{R}\right|}\sum_{(u,i)\in
  O_R}\left(\hat{r}_{u,i}-r_{u,i}\right)^{2},
\end{align}
where
$\TT=\left\{ \mu, \left\{ \beta_u\right\}_{u=1}^N, \left\{ \beta_i
  \right\}_{i=1}^M, \left\{ \gamma_u\right\}_{u=1}^N, \left\{
    \vgamma_i\right\}_{i=1}^M \right\}$.

Once the parameters $\TT$ are estimated by minimizing $C_R$, it is
straightforward to predict the rating of an unobserved user-product
pair $(u,i)$ using Eq.~\eqref{eq:matfact}.

\subsection{Taming the Curse of Data Sparsity}

It has been observed earlier, for instance in \cite{Ilin2010}, that
this matrix factorization approach easily overfits the observed
ratings, leading to poor generalization performance on the held-out
set, or unseen user-product pairs. This issue is especially serious in
the case of recommendation systems, as it is highly likely that each
user purchases/watches only a fraction of all the available products.
For instance, in the Amazon Reviews Dataset more than 99.999\% of
ratings, or elements in the rating matrix are missing.

The issue of overfitting is often addressed by adding a regularization
term $\Omega$ to the cost $C_R$ in Eq.~\eqref{eq:cost_r}. One of the
most widely used regularization term is a simple weight decay
\begin{align*}
\Omega(\TT) = \sum_{\theta \in \TT} \left\| \theta \right\|^2.
\end{align*}
Hence parameters are estimated by minimizing
$C_R(\TT) + \lambda \Omega(\TT)$, where $\lambda$ is a regularization
coefficient.

Another approach is to interpret matrix factorization in a
probabilistic
framework~\cite{Ilin2010,mnih2007:pmf,salakhutdinov2008:bpmf}. In this
approach, all the parameters such as the user and product
representations are considered as latent random variables on which the
distribution of the rating, an observed random variable, is
conditioned. This probabilistic matrix factorization can automatically
regularize itself by maintaining the confidence of the
estimates/predictions based on the observations.

On the other hand, we can improve generalization, hence reduce
overfitting, by simultaneously estimating the parameters $\TT$ of the
matrix factorization to perform well on another related
task~\cite{caruana1997multitask}. With a model predicting a rating by
a user on a product, we can consider letting the model also try to
account for the product review given by the user. This seems like a
useful side task as users often write reviews that justify their
ratings and describe features that affected their opinions.

In this paper, we explore this approach of exploiting extra tasks to
improve generalization performance of collaborative filtering based on
matrix factorization.

\section{Regularizing with Extra Data}

\subsection{Reviews as Extra Data}

In many e-commerce systems, each rating is often accompanied with a
user's review of the product. As mentioned earlier, it is natural to
expect that the accompanying review is used by the user to justify
her/his rating or opinion, which suggests the possibility of improving
the generalization performance of the prediction model for
ratings~\cite{McAuley13:hft}.  As an illustrating example, a user, who
wrote ``this is a great adventure movie that children and adults alike
would love!'' for the movie ``Free Willy'', is likely to give a higher
rating to this movie.\footnote{This is an actual sample from the
  Amazon Reviews datatset.}

In this section, we will propose two approaches to utilizing this type
of review data for improving the generalization performance of a
rating prediction model based on matrix factorization.

\subsection{Natural Language Review Modeling}

More technically, let us suppose that the set $R$ of ratings (see
Sec.~\ref{sec:mf}) is accompanied with a set $D$ of reviews
$D=\left\{ d_{u,i}\right\}_{(u,i)\in O_D}$. Each review
$d_{u,i}=\left(w_{u,i}^{(1)}, \cdots, w_{u,i}^{(n_{u,i})}\right)$ is a
piece of natural language text written by a user $u$ about an item
$i$, which we represent as a sequence of words.

Following the multitask learning
framework~\cite{caruana1997multitask}, we build a model that jointly
predicts the rating given by a user $u$ to a product $i$ and models
the review written by the user $u$ on the product $i$. The model has
two components; matrix factorization in Eq.~\eqref{eq:matfact} and
review modeling, which shares some of the parameters $\TT$ from the
rating prediction model.

Here, we follow the approach from \cite{McAuley13:hft} by modeling the
conditional probability of each review given the corresponding product
$\vgamma_i$:
\begin{align}
  \label{eq:review_model}
  p\left(d_{u,i}=\left(w_{u,i}^{(1)}, \cdots, w_{u,i}^{(n_{u,i})}\right)\mid
  \vgamma_i, \TT_D \right),
\end{align}
where $\TT_D$ is a set of parameters for this review model.

We estimate the parameters of this review model ($\TT_D$ and
$\vgamma_i$'s) by minimizing the negative log-likelihood:
\begin{align*}
  \argmin_{\TT_D, \left\{\vgamma_i\right\}_{i=1}^M} C_D(\TT_D,
  \left\{\vgamma_i\right\}_{i=1}^M),
\end{align*}
where
\begin{align}
  \label{eq:cost_d}
  C_D(&\TT_D, \left\{\vgamma_i\right\}_{i=1}^M) = \\
      &-\frac{1}{\left| O_D \right|}
        \sum_{(u,i)\in O_D} \log p\left(d_{u,i}=\left(w_{u,i}^{(1)}, \cdots,
        w_{u,i}^{(n_{u,i})}\right) \mid \vgamma_i\right).
\end{align}

We jointly optimize the rating prediction model in
Eq.~\eqref{eq:matfact} and the review model in
Eq.~\eqref{eq:review_model} by minimizing the convex combination of
$C_R$ in Eq.~\eqref{eq:cost_r} and $C_D$ in Eq.~\eqref{eq:cost_d}:
\begin{align}
  \label{eq:cost_joint}
  \argmin_{\TT, \TT_D} \alpha\,C_R(\TT) + (1-\alpha) C_D(\TT_D, \left\{ \vgamma_i
  \right\}_{i=1}^M),
\end{align}
where the coefficient $\alpha$ is a hyperparmeter.

\subsubsection{BoWLF: Distributed Bag-of-Word}
\label{sec:bowlf}

The first model we propose to use is a distributed bag-of-words
prediction. In this case, we represent each review as a bag of words,
meaning
\begin{align}
  \label{eq:bow}
  d_{u,i} = 
  \left( w_{u,i}^{(1)}, \cdots, w_{u,i}^{(n_{u,i})} \right) \approx
  \left\{ w_{u,i}^{(1)}, \cdots, w_{u,i}^{(n_{u,i})} \right\}. 
\end{align}
This leads to
\begin{align*}
  p(d_{u,i} \mid \vgamma_i) = 
  \prod_{t=1}^{n_{u,i}} p(w_{u,i}^{(t)} \mid \vgamma_i).
\end{align*}

We model $p(w_{u,i}^{(t)} \mid \vgamma_i)$ as an affine transformation
of the product representation $\vgamma_i$ followed by, so-called
softmax, normalization:
\begin{align}
  \label{eq:bowlf}
  p(w_{u,i}^{(t)}=j\mid \vgamma_i) = \frac{\exp\left\{ y_j \right\}}{
  \sum_{l=1}^{|V|} \exp\left\{ y_l \right\}},
\end{align}
where
\begin{align*}
  \vy = \mW \vgamma_i+ \vb
\end{align*}and $V$, $\mW$ and $\vb$ are the vocabulary, a weight matrix and a bias vector.
The parameters $\TT_D$ of this review model include $\mW$ and $\vb$.

When we use this distributed bag-of-words together with matrix
factorization for predicting ratings, and we call this joint model the
{\it bag-of-words regularized latent factor model} (BoWLF).

\subsubsection{LMLF: Recurrent Neural Network}
\label{sec:lmlf}

The second model of reviews we propose to use is a recurrent neural
network (RNN) language model (LM)~\cite{Mikolov-thesis-2012}. Unlike
the distributed bag-of-words model, this RNN-LM does not make any
assumption on how each review is represented, but takes a sequence of
words as it is, preserving the order of the words.

In this case, we model the probability over a review which is a
variable-length sequence of words by rewriting the probability as
\begin{align*}
  p(&d_{u,i}=( w_{u,i}^{(1)}, \cdots, w_{u,i}^{(n_{u,i})}) \mid \vgamma_i) \\ =&
                                                                                 p\left(w_{u,i}^{(1)} \mid \vgamma_i\right) \prod_{t=2}^{n_{u,i}} p\left(w_{u,i}^{(t)} \mid w_{u,i}^{(1)},
                                                                                 \cdots, w_{u,i}^{(t-1)}, \vgamma_i\right),
\end{align*}

We approximate each conditional distribution with
\begin{align*}
  p\left(w_{u,i}^{(t)} = j \mid w_{u,i}^{(<t)}, \vgamma_i\right) = \frac{\exp\left\{ y_j^{(t)} \right\}}{
  \sum_{l=1}^{|V|} \exp\left\{ y_l^{(t)} \right\}},
\end{align*}
where
\begin{align*}
  \vy^{(t)} = \mW \vh^{(t)} + \vb
\end{align*}
and
\begin{align*}
  \vh^{(t)} = \phi\left( \vh^{(t-1)}, w_{u,i}^{(t-1)}, \vgamma_i \right).
\end{align*}

There are a number of choices available for implementing the recurrent
function $\phi$. Here, we use a long short-term memory
(LSTM,~\cite{Hochreiter+Schmidhuber-1997}) which has recently been
applied successfully to natural language-related
tasks~\cite{Graves-arxiv2013}.

In the case of the LSTM, the recurrent function $\phi$ returns, in
addition to its hidden state $\vh^{(t)}$, the memory cell $\vc^{(t)}$
such that
\begin{align*}
  \left[ \vh^{(t)}; \vc^{(t)}\right] = \phi\left( \vh^{(t-1)}, \vc^{(t-1)}, w_{u,i}^{(t-1)}, \vgamma_i \right),
\end{align*}
where
\begin{align*}
  &\vh^{(t)} = \vo^{(t)} \odot \tanh(\vc^{(t)}) \\
  &\vc^{(t)} = \vf^{(t)} \odot \vc^{(t-1)} + \vi^{(t)} \odot
    \tilde{\vc}^{(t)}.
\end{align*}
The output $\vo$, forget $\vf$ and input $\vi$ gates are computed by
\begin{multline}
  \left[\begin{array}{c}
          \vo^{(t)} \\
          \vf^{(t)} \\
          \vi^{(t)}
        \end{array}\right]
  = 
  \sigma ( \mV_g \mE\left[ w_{u,i}^{(t-1)} \right] +  
  \mW_g \vh^{(t-1)} + \\ 
  \mU_g \vc^{(t-1)} +
  \mA_g \vgamma_i + \vb_g ),
\end{multline}
and the new memory content $\tilde{\vc}^{(t)}$ by
\begin{multline}
  \tilde{\vc}^{(t)} = \tanh( \mV_c \mE\left[ w_{u,i}^{(t-1)} \right] + 
  \mW_c \vh^{(t-1)} + \\ 
  \mU_c \vc^{(t-1)} + \mA_c \vgamma_i + \vb_c),
\end{multline}
where $\mE$, $\mV_g$, $\mW_g$, $\mU_g$, $\vb_g$, $\mV_c$, $\mW_c$,
$\mU_c$, $\vb_c$, $\mA_g$ and $\mA_c$ are the parameters of the
RNN-LM. Note that $\mE\left[ w \right]$ denotes a row indexing by the
word index $w$ of the matrix $\mE$.

Similarly to the BoWLF, we call the joint model of matrix
factorization and this RNN-LM the {\it language model regularized
  latent factor model} (LMLF).

\subsection{Related Work: LDA-based Approach}
\label{sec:hft}

Similar approaches of modeling reviews to regularize matrix
factorization have recently been proposed, however, with different
review models such as LDA~\cite{McAuley13:hft,Ling2014:RMR} and
non-negative matrix factorization~\cite{bao2014topicmf}. Here, we
describe ``Hidden Factors as Topics'' (HFT) recently proposed in
\cite{McAuley13:hft}, and discuss it with respect to the proposed
approaches.

The HFT model is based on latent Dirichlet allocation
(LDA,~\cite{blei2003latent}), and similarly to the distributed
bag-of-word model in Sec.~\ref{sec:bowlf}, considers each review as a
bag of words (see Eq.~\eqref{eq:bow}.) Thus, we start by describing
how LDA models a review $d_{u,i}$.

LDA is a generative model of a review/document. It starts by sampling
a so-called topic proportion $\vtau$ from a Dirichlet
distribution. $\vtau$ is used as a parameter to a multinomial topic
distribution from which a topic is sampled. The sampled topic defines
a probability distribution over the words in a vocabulary. In other
words, given a topic proportion, the LDA models a review with a
mixture of multinomial distributions.

Instead of sampling the topic proportion from the top-level Dirichlet
distribution in LDA, HFT replaces it with
\[
  \vtau = \frac{1}{\left\| \exp\left\{ \kappa \vgamma_i
      \right\}\right\|_1} \exp\left\{ \kappa \vgamma_i\right\},
\]
where $\kappa$ is a free parameter estimated along with all the other
parameters of the model.  In this case, the probability over a single
review $d_{u,i}$ given a product $\vgamma_i$ becomes
\begin{align}
  p(d_{u,i} \mid \vgamma_i) =& 
                               \prod_{t=1}^{n_{u,i}} \sum_{k=1}^{\text{dim}(\vgamma_i)}
                               p(w_{u,i}^{(t)} \mid z_{k}=1) p(z_{k}=1 \mid \vgamma_i)  
                               \label{eq:hft_mixture} \\
  =& \prod_{t=1}^{n_{u,i}} \sum_{k=1}^{\text{dim}(\vgamma_i)}
     \tau_k p(w_{u,i}^{(t)} \mid z_{k}=1) \nonumber 
\end{align}
where $z_k$ is an indicator variable of the $k$-th topic out of
$\text{dim}(\vgamma_i)$, and $\tau_{k}$ is the $k$-th element of
$\vtau$. The conditional probability over words given a topic is
modeled with a stochastic matrix
$\mW^* = \left[ w^*_{j,k} \right]_{|V| \times \text{dim}(\vgamma_i)}$
(each column sums to $1$).  The conditional probability over words
given a product $\vgamma_i$ can be written as
\begin{align}
  \label{eq:hft}
  p(w_{u,i}^{(t)}=j \mid \vgamma_i) = \sum_{k=1}^{\text{dim}(\vgamma_i)} 
  w^*_{j,k} \frac{\exp\left\{ \kappa \gamma_{i,k}
  \right\}}{\left\| \exp\left\{ \kappa \vgamma_i \right\}\right\|_1}.
\end{align}
The matrix $\mW^*$ is often parametrized by
$w^*_{j,k} = \frac{\exp\left\{ q_{j,k}\right\}}{\sum_{l} \exp\left\{
    q_{l,k}\right\}}$,
where $\mQ=\left[q_{j,k}\right]$ is an unconstrained matrix of the
same size as $\mW^*$.  In practice, a bias term is added to the
formulation above to handle frequent words.

\subsection{Comparing HFT and BoWLF}
\label{sec:compare}
From Eq.~\eqref{eq:bowlf} and Eq.~\eqref{eq:hft}, we can see that the
HFT and the proposed BoWLF (see Sec.~\ref{sec:bowlf}) are closely
related. Most importantly, both of them consider a review as a bag of
words and parametrize the conditional probability of a word given a
product representation with a single affine transformation (weight
matrix plus offset vector).

The main difference is in how the product representation and the weight
matrix interact to form a point on the $|V|$-dimensional simplex.  In
the case of HFT, both the product representation $\vgamma_i$ and the
projection matrix $\mW^*$ are separately stochastic (i.e. each
$\vgamma_i$ and each column of $\mW^*$ are interpretable as a
probability distribution), while the BoWLF projects the result of the
matrix-vector product $\mW \vgamma_i$ onto the probability simplex.

This can be understood as the difference between a mixture of experts
and a product of experts~\cite{Hinton99}. On a per word basis, the
BoWLF in Eq.~\eqref{eq:bowlf} can be re-written as a (conditional)
product of experts by
\begin{align*}
  p(w=j \mid \vgamma_i) = \frac{1}{Z(\vgamma_i)}
  \prod_{k=1}^{\text{dim}(\vgamma_i)} \exp\left\{ w_{j,k} \gamma_{i,k} + b_j
  \right\},
\end{align*}
where $w_{j,k}$ and $b_j$ are the element at the $j$-th row and $k$-th
column of $\mW$ and the $j$-th element of $\vb$, respectively.  On the
other hand, an inspection of Eq.~\eqref{eq:hft_mixture} reveals that,
on a per word basis, the HFT model is clearly a mixture model, with
the topics playing the role of the mixture components.

As argued in \cite{Hinton99}, a product of experts can more easily
model a {\it peaky} distribution, especially, in a high-dimensional
space.\footnote{ Note that the size of a usual vocabulary of reviews
  is on the order of thousands. }  The reviews of each product tend to
contain a small common subset of the whole vocabulary, while those
subsets vastly differ from each other depending on the product. In
other words, the conditional distribution of words given a product
puts most of its probability mass on only a few product-specific
words, while leaving most other words with nearly zero
probabilities. Product of experts are naturally better suited to
modeling peaky distributions rather than mixture models.

A more concrete way of understanding the difference between HFT and
BoWLF may be to consider how the product representation and the weight
matrix interact. In the case of the BoWLF, this is a simple
matrix-vector product with no restrictions on the weight matrix. This
means that both the product representation elements as well as the
elements of the weight matrix are free to assume negative
values. Thus, it is possible that an element of the product
representation could exercise a strong influence \emph{suppressing}
the expression of a given set of words. Alternatively, with HFT model,
as the model interprets the elements of the product representation as
mixture components, these elements have no mechanism of suppressing
probability mass assigned to words by the other elements of the
product representation.

We suggest that this difference allows the BoWLF to better model
reviews compared to the HFT, or any other LDA-based model by offering
a mechanism for negative correlations between words to be explicitly
expressed by elements of the product representation. By offering a
more flexible and natural model of reviews, the BoWLF model can
improve the rating prediction generalization performance. As we will
see in Sec.~\ref{sec:exps}, our experimental results support this
proposition.

The proposed LMLF takes one step further by modeling each review with
a chain of products of experts taking into account the order of
words. This may seem an obvious benefit at the first sight. However,
it is not clear whether the order of the words is specific to a
product or is simply a feature of language itself.  In the latter
case, we expect that LMLF will model reviews very well, but may not
improve rating prediction.

\section{Experiments}
\label{sec:exps}
\subsection{Dataset}
We evaluate the proposed approaches on the Amazon Reviews
dataset~\cite{McAuley13:hft}.\footnote{https://snap.stanford.edu/data/web-Amazon.html}
There are approximate 35 million ratings and accompanying reviews from
6,643,669 users and 2,441,053 products. The products are divided into
28 categories such as music and books. The reviews are on average 110
words long. We refer the reader to \cite{Ling2014:RMR} for more
detailed statistics.

\subsection{Experimental Setup}

\paragraph{Data Preparation}
We closely follow the procedure from \cite{McAuley13:hft} and
\cite{Ling2014:RMR}, where the evaluation is done per category.  We
randomly select 80\% of ratings, up to two million samples, as a
training set, and split the rest evenly into validation and test sets,
for each category. We preprocess reviews only by tokenizing them using
a script from Moses\footnote{
  \url{https://github.com/moses-smt/mosesdecoder/} }, after which we
build a vocabulary of 5000 most frequent words.

\paragraph{Evaluation Criteria}
We use mean squared error (MSE) of the rating prediction to evaluate
each approach. For assessing the performance on review modeling, we
use the average negative log-likelihood.

\paragraph{Baseline}
We compare the two proposed approaches, BoWLF (see
Sec.~\ref{sec:bowlf}) and LMLF (see Sec.~\ref{sec:lmlf}), against
three baseline methods; matrix factorization with $L_2$ regularization
(MF, see Eqs.~\eqref{eq:matfact}--\eqref{eq:cost_r}), the HFT model
from \cite{McAuley13:hft} (see Sec.~\ref{sec:hft}) and the RMR model
from \cite{Ling2014:RMR}. In the case of HFT, we report the
performance both by evaluating the model ourselves\footnote{ The code
  was kindly provided by the authors of \cite{McAuley13:hft}.  } and
by reporting the results from \cite{McAuley13:hft} directly. For RMR,
we only report the results from \cite{Ling2014:RMR}.


\paragraph{Hyper-parameters}
Both user $\vgamma_u$ and product $\vgamma_i$ vectors in
Eq.~\eqref{eq:matfact} are five dimensional for all the experiments in
this section. This choice was made mainly to make the results
comparable to the previously reported ones in \cite{McAuley13:hft} and
\cite{Ling2014:RMR}.

We initialize all the user and product representations by sampling
each element from a zero-mean Gaussian distribution with its standard
deviation set to 0.01. The biases, $\mu$, $\beta_u$ and $\beta_i$ are
all initialized to 0. All the parameters in BoWLF and LMLF are
initialized similarly except for the recurrent weights of the RNN-LM
in LMLF which were initialized to be orthogonal.

\paragraph{Training Procedure}
When training MF, BoWLF and LMLF, we use minibatch RMSProp with the
learning rate, momentum coefficient and the size of minibatch set to
$0.01$, $0.9$ and $128$, respectively. We trained each model at most
$200$ epochs, while monitoring the validation performance. For HFT, we
follow \cite{McAuley13:hft} which uses the Expectation Maximization
algorithm together with L-BFGS. In all cases, we early-stop each
training run based on the validation set performance.

In the preliminary experiments, we found the choice of $\alpha$ in
Eq.~\eqref{eq:cost_joint}, which balances matrix factorization and
review modeling, to be important. We searched for the $\alpha$ that
maximizes the validation performance, in the range of
$\left[0.1, 0.01\right]$.

We used a CPU cluster of $16$ nodes each with $8$ cores and $8-16$ GB
of memory to run experiments on BoWLF, MF, and HFT. For LMLF, we used
a cluster of K20 GPUs where we had up to 50 GPUs available.
  
\subsection{Rating Prediction Results}
\label{sec:quant}

\begin{table*}[!ht]\ \small

\begin{tabular}{p{20mm}|p{10mm}llllll||ll}
  \toprule
  &    Dataset      & (a) & (b) & (c)   & (d)  &
                                                 \multicolumn{2}{l||}{BoWLF improvement} &       &       \\
  Dataset     &    Size  & MF  & HFT & BoWLF & LMLF & over (a)           & over (b)           & HFT*  & RMR** \\
  \midrule
  Arts                     & 27K      & 1.434 (0.04)     & 1.425 (0.04)     & \textbf{1.413} (0.04)      & 1.426 (0.04)     & 2.15\%      & 1.18\%       & 1.388 & 1.371 \\
  Jewelry                  & 58K      & 1.227 (0.04)     & \textbf{1.208} (0.03)     & 1.214 (0.03)      & 1.218 (0.03)     & 1.24\%      & -0.59\%      & 1.178 & 1.160 \\
  Watches                  & 68K      & 1.511 (0.03)     & 1.468 (0.03)     & \textbf{1.466} (0.03)      & 1.473 (0.03)     & 4.52\%      & 0.20\%       & 1.486 & 1.458 \\
  Cell Phones              & 78K      & 2.133 (0.03)     & 2.082 (0.02)     & \textbf{2.076} (0.02)      & 2.077 (0.02)     & 5.76\%      & 0.66\%       & N/A   & 2.085 \\
  Musical Inst.            & 85K      & 1.426 (0.02)     & 1.382 (0.02)     & \textbf{1.375} (0.02)      & 1.388 (0.02)     & 5.12\%      & 0.75\%       & 1.396 & 1.374 \\
  Software                 & 95K      & 2.241 (0.02)     & 2.194 (0.02)     & \textbf{2.174} (0.02)      & 2.203 (0.02)     & 6.70\%      & 2.06\%       & 2.197 & 2.173 \\
  Industrial               & 137K     & 0.360 (0.01)     & 0.354 (0.01)     & \textbf{0.352} (0.01)      & 0.356 (0.01)     & 0.76\%      & 0.24\%       & 0.357 & 0.362 \\
  Office Products          & 138K     & 1.662 (0.02)     & 1.656 (0.02)     & \textbf{1.629} (0.02)      & 1.646 (0.02)     & 3.32\%      & 2.72\%       & 1.680 & 1.638 \\
  Gourmet Foods            & 154K     & 1.517 (0.02)     & 1.486 (0.02)     & \textbf{1.464} (0.02)      & 1.478 (0.02)     & 5.36\%      & 2.22\%       & 1.431 & 1.465 \\
  Automotive               & 188K     & 1.460 (0.01)     & 1.429 (0.01)     & \textbf{1.419} (0.01)      & 1.428 (0.01)     & 4.17\%      & 1.03\%       & 1.428 & 1.403 \\
  Kindle Store             & 160K     & 1.496 (0.01)     & 1.435 (0.01)     & \textbf{1.418} (0.01)      & 1.437 (0.01)     & 7.83\%      & 1.76\%       & N/A   & 1.412 \\
  Baby                     & 184K     & 1.492 (0.01)     & 1.437 (0.01)     & \textbf{1.432} (0.01)      & 1.443 (0.01)     & 5.95\%      & 0.48\%       & 1.442 & N/A    \\
  Patio                    & 206K     & 1.725 (0.01)     & 1.687 (0.01)     & \textbf{1.674} (0.01)      & 1.680 (0.01)     & 5.10\%      & 1.24\%       & N/A   & 1.669 \\
  Pet Supplies             & 217K     & 1.583 (0.01)     & 1.554 (0.01)     & \textbf{1.536} (0.01)      & 1.544 (0.01)     & 4.74\%      & 1.78\%       & 1.582 & 1.562 \\
  Beauty                   & 252K     & 1.378 (0.01)     & 1.373 (0.01)     & \textbf{1.335} (0.01)      & 1.370 (0.01)     & 4.33\%      & 3.82\%       & 1.347 & 1.334 \\
  Shoes                    & 389K     & 0.226 (0.00)     & 0.231 (0.00)     & \textbf{0.224} (0.00)      & 0.225 (0.00)     & 0.23\%      & 0.72\%       & 0.226 & 0.251 \\
  Tools \& Home            & 409K     & 1.535 (0.01)     & 1.498 (0.01)     & \textbf{1.477} (0.01)      & 1.490 (0.01)     & 5.78\%      & 2.15\%       & 1.499 & 1.491 \\
  Health                   & 428K     & 1.535 (0.01)     & 1.509 (0.01)     & \textbf{1.481} (0.01)      & 1.499 (0.01)     & 5.35\%      & 2.82\%       & 1.528 & 1.512 \\
  Toys \& Games            & 435K     & 1.411 (0.01)     & 1.372 (0.01)     & \textbf{1.363} (0.01)      & 1.367 (0.01)     & 4.71\%      & 0.89\%       & 1.366 & 1.372 \\
  Video Games              & 463K     & 1.566 (0.01)     & 1.501 (0.01)     & \textbf{1.481} (0.01)      & 1.490 (0.01)     & 8.47\%      & 2.00\%       & 1.511 & 1.510 \\
  Sports                   & 510K     & 1.144 (0.01)     & 1.137 (0.01)     & \textbf{1.115} (0.01)      & 1.127 (0.01)     & 2.94\%      & 2.19\%       & 1.136 & 1.129 \\
  Clothing                 & 581K     & 0.339 (0.00)     & 0.343 (0.00)     & \textbf{0.333} (0.00)      & 0.344 (0.00)     & 0.60\%      & 1.01\%       & 0.327 & 0.336 \\
  Amazon Video             & 717K     & 1.317 (0.01)     & 1.239 (0.01)     &
                                                                              \textbf{1.184} (0.01)      & 1.206 (0.01)     & 13.33\%      & 5.47\%       & N/A   & 1.270 \\
  Home                     & 991K     & 1.587 (0.00)     & 1.541 (0.00)     & \textbf{1.513} (0.00)      & 1.535 (0.01)     & 7.41\%      & 2.79\%       & 1.527 & 1.501 \\
  Electronics              & 1.2M     & 1.754 (0.00)     & 1.694 (0.00)     & \textbf{1.671} (0.00)      & 1.698 (0.00)     & 8.29\%      & 2.30\%       & 1.724 & 1.722 \\
  Music                    & 6.3M     & 1.112 (0.00)     & 0.970 (0.00)     & \textbf{0.920} (0.00)      & 0.924 (0.00)     & 19.15\%      & 4.94\%       & 0.969 & 0.959 \\
  Movies \& Tv             & 7.8M     & 1.379 (0.00)     & 1.089 (0.00)     & \textbf{0.999} (0.00)      & 1.022 (0.00)     & 37.95\%      & 9.01\%       & 1.119 & 1.120 \\
  Books                    & 12.8M    & 1.272 (0.00)     & 1.141 (0.00)     & \textbf{1.080} (0.00)      & 1.110 (0.00)     & 19.21\%      & 6.12\%       & 1.135 & 1.113 \\
  \midrule
  All categories           & 35.3M    & 1.289            & 1.143            & 1.086                    & 1.107                   & 20.29\%            & 5.64\%             &       &  \\    
  \bottomrule
\end{tabular}
\caption{Prediction Mean Squared Error results on test data. Standard error of
  mean in parenthesis. Dimensionality of latent factors $\text{dim}(\vgamma_i)=5$ for all
  models. Best results for each dataset in bold. HFT* and RMR** represent original paper
  results over different data splits \cite{McAuley13:hft,Ling2014:RMR}.}
\label{table:results}
\end{table*}

\begin{figure}[ht]
  \includegraphics[width=.95\columnwidth]{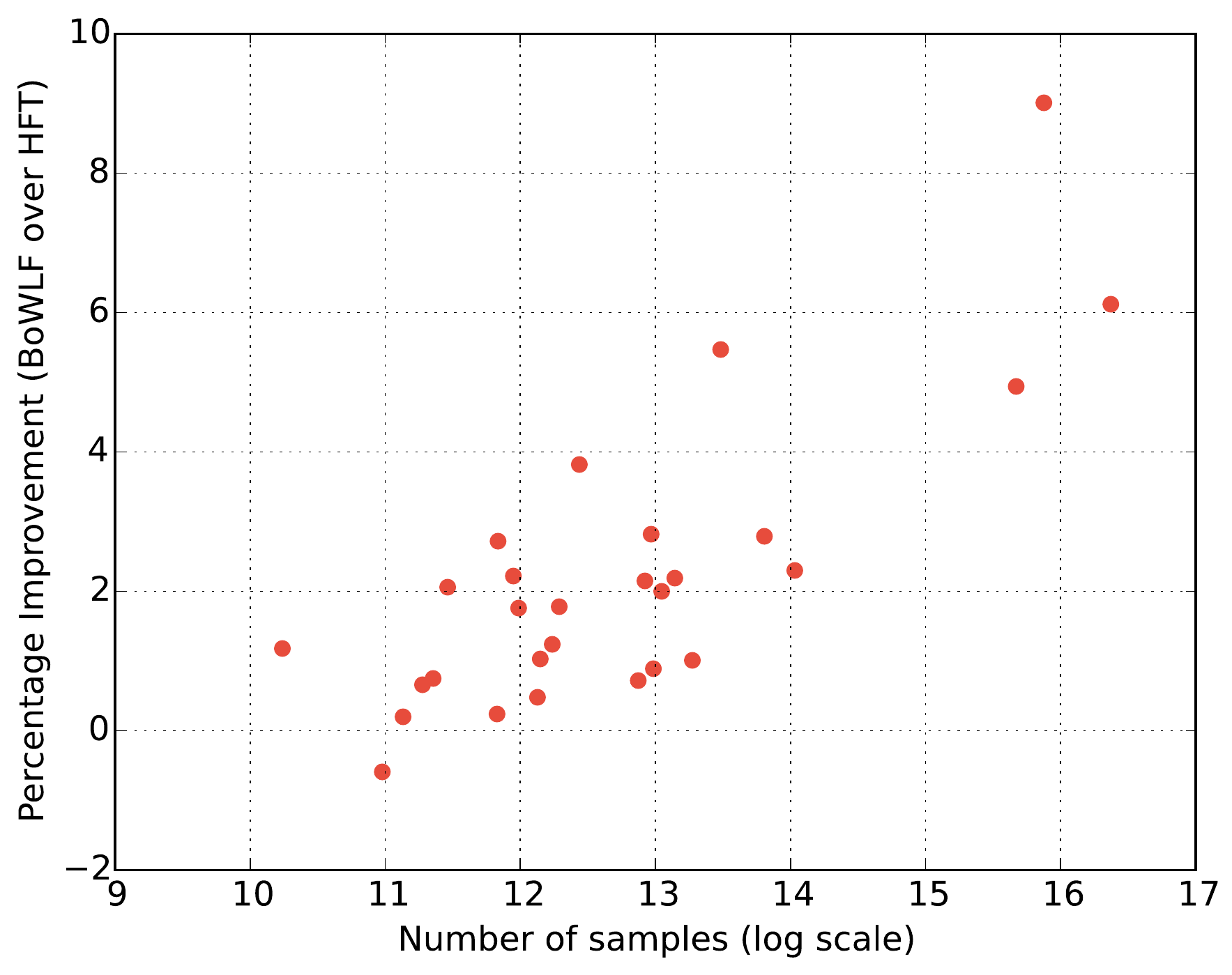}
  \caption{Scatterplot showing performance improvement over the number
    of samples.  We see a performance improvement of BoWLF over HFT as
    dataset size increases.}
  \label{figure:scatterplot_bowlf}
\end{figure}

\begin{figure}[ht]
  \includegraphics[width=.95\columnwidth]{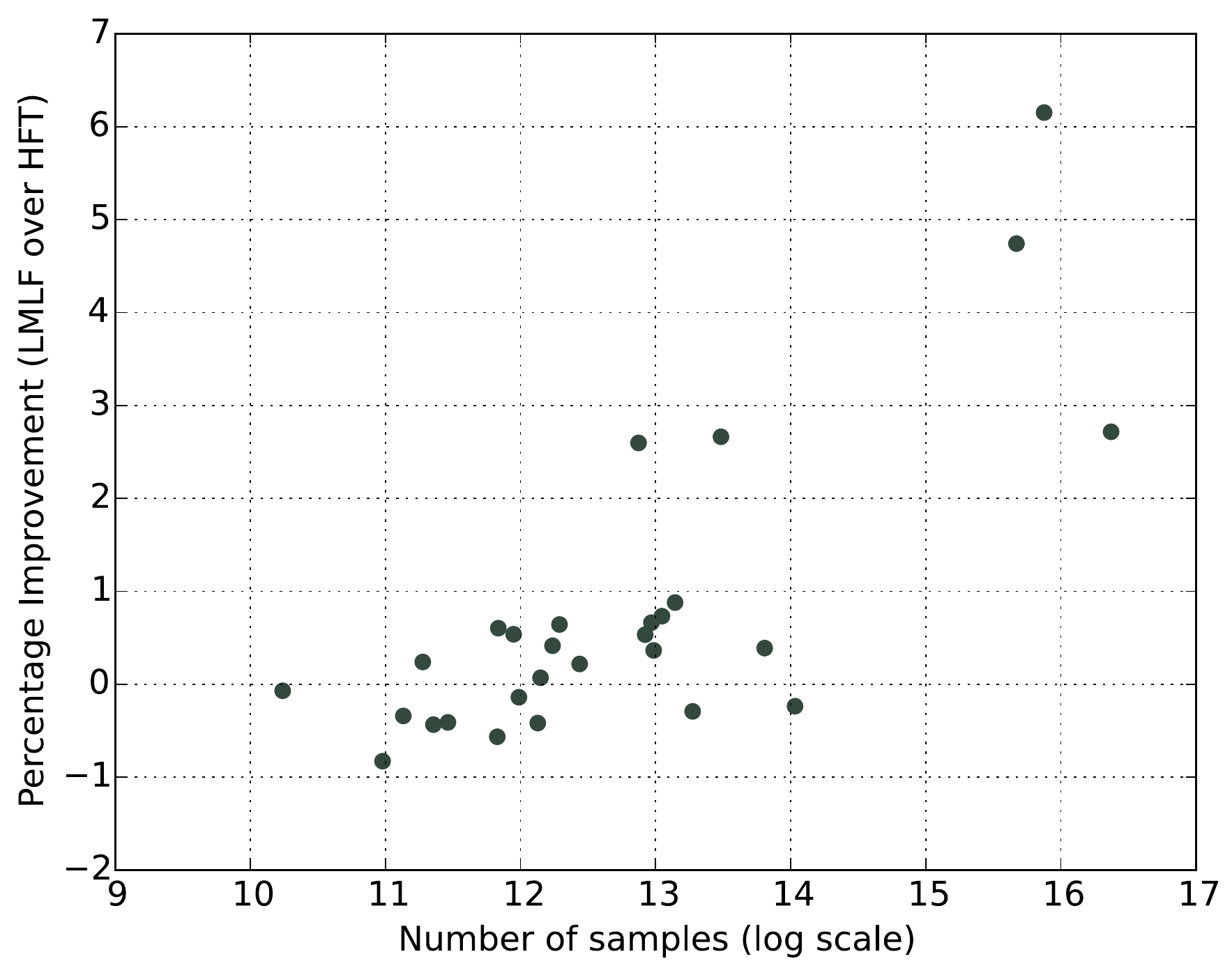}
  \caption{Scatterplot showing performance improvement over the number
    of samples.  We see a modest performance improvement of LMLF over
    HFT as dataset size increases.}
  \label{figure:scatterplot_lmlf}
\end{figure}

We list results of the experiments in Table~\ref{table:results} for
the 28 categories in terms of MSE with the standard error of mean
shown in parentheses. From this table, we can see that except for a
single category of ``Jewelry'', the proposed BoWLF outperforms all the
other models with an improvement of 20.29\% over MF and 5.64\% over
HFT across all categories.\footnote{ Due to the use of different
  splits, the results by HFT reported in \cite{McAuley13:hft} and RMR
  in \cite{Ling2014:RMR} are not directly comparable.  }  In general,
we note better performance of BoWLF and LMLF models over other methods
especially as the size of the dataset grows, which is evident from
Figs. ~\ref{figure:scatterplot_bowlf} and
~\ref{figure:scatterplot_lmlf}.



\newpage
Interestingly, BoWLF always outperforms LMLF. These results
indicate that the complex language model, which the LMLF learns using
an LSTM network, does not seem to improve over a simple bag-of-word
representation, which the BoWLF learns, in terms of the learned
product representations.

This can be understood from how the product representation, which is
used {\it linearly} by the rating prediction model, is handled by each
model. The word distribution modeled by the BoWLF depends linearly on
the product representation, which requires the product-related
structure underlying reviews be encoded linearly as well. On the other
hand, LMLF {\it nonlinearly} manipulates the product representation to
approximate the distribution over reviews. In other words, the LMLF
does not necessarily encode the underlying product-related structure
inside the product representation in the way the rating prediction
model can easily decode~\cite{Mikolov-et-al-ICLR2013}.

\begin{figure}
  \advance\leftskip-.02\columnwidth
  \makebox[\columnwidth][c]{\includegraphics[height=.36\textheight,
    width=\columnwidth]{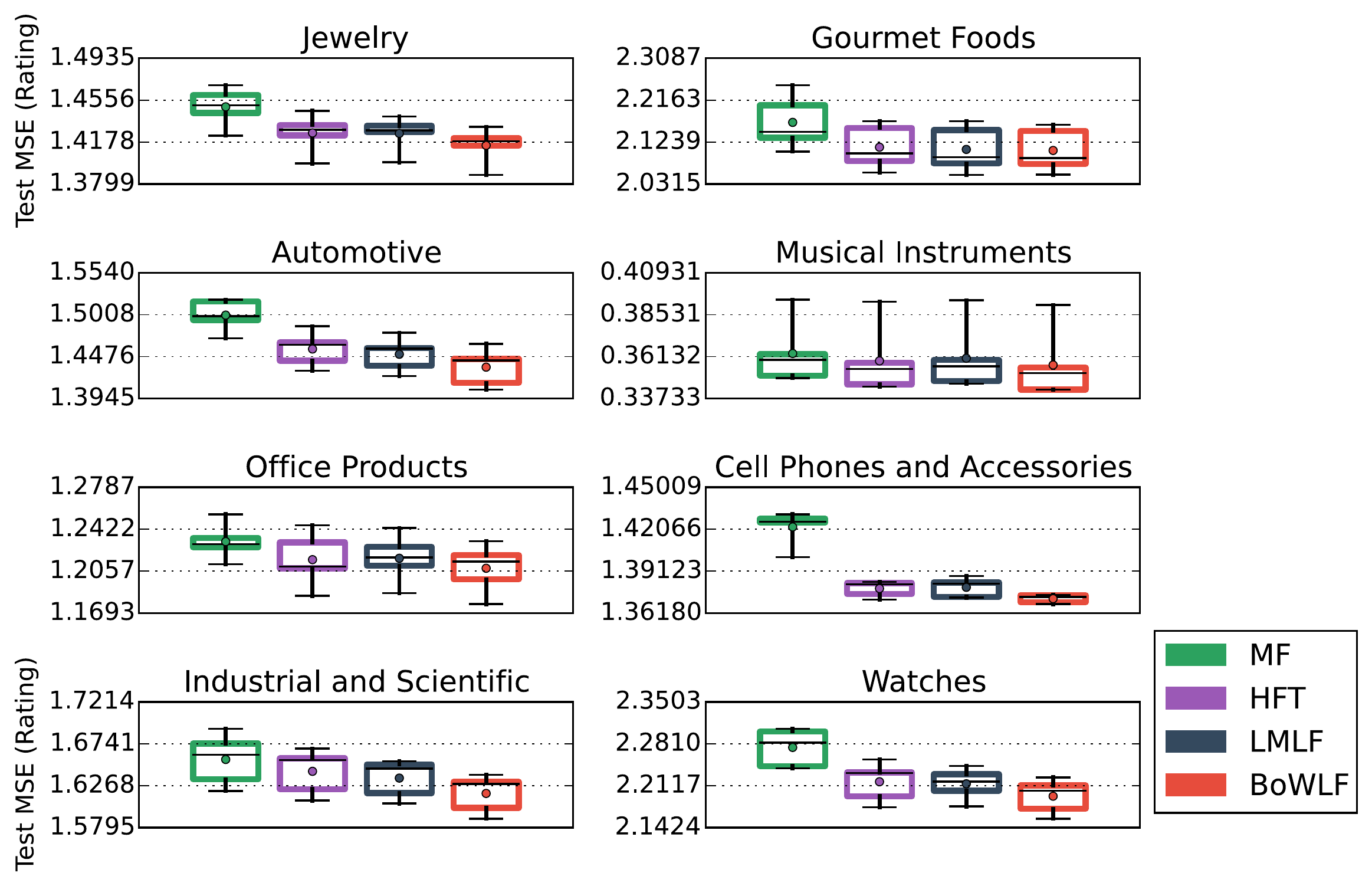}}
  \caption{Box and whisker plot showing K-fold ($K=5$) experiments.
    Point represents the mean over all folds. Center line represents
    median. Box extents represent $25^{\text{th}}$ and
    $75^{\text{th}}$ percentile. Whisker extents show minimum and
    maximum values.}
  \label{figure:boxplot}
\end{figure}

\begin{figure}
  \advance\rightskip-.02\columnwidth
  \makebox[\columnwidth][c]{\includegraphics[height=.36\textheight,
    width=1.1\columnwidth]{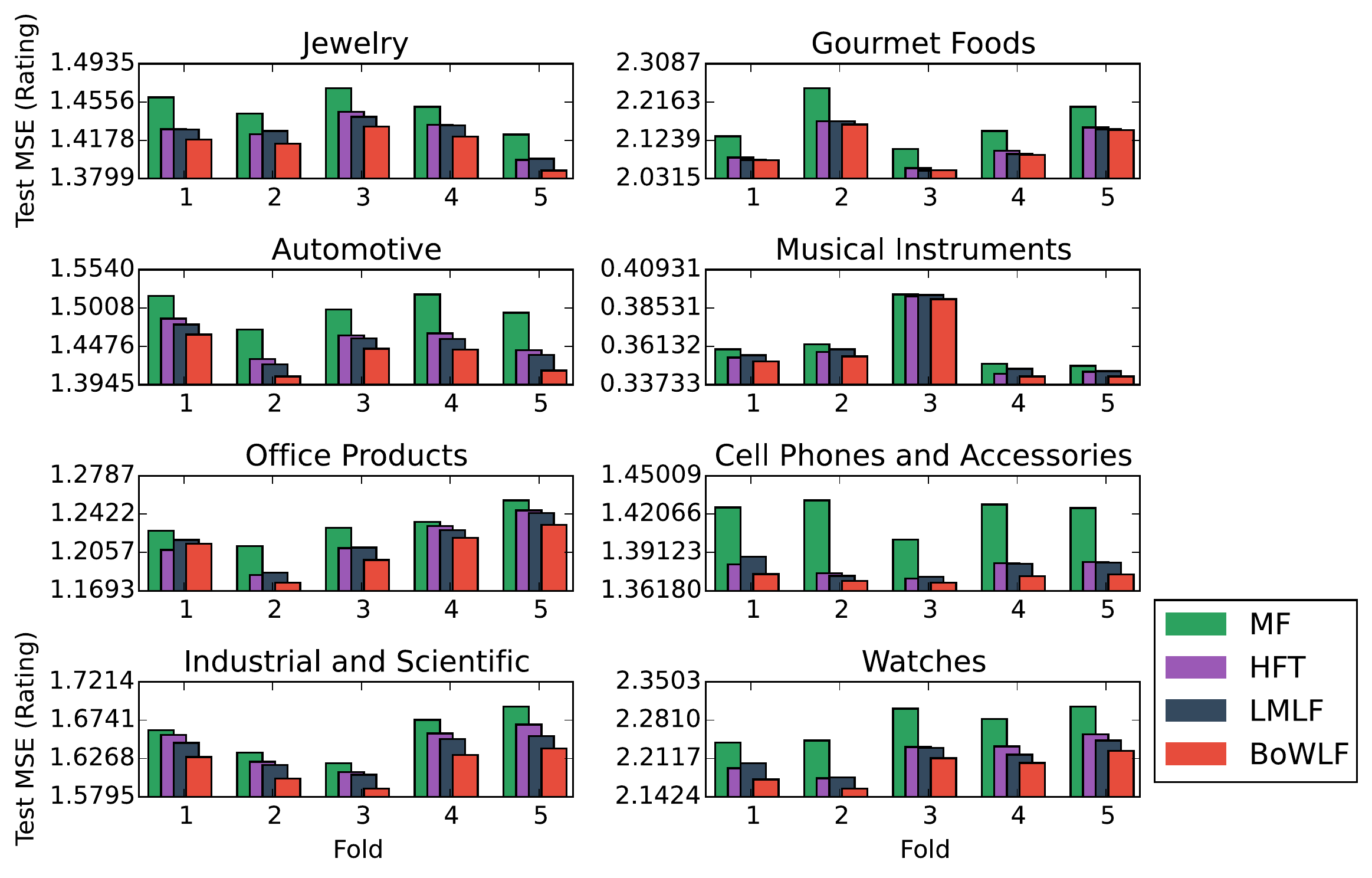}}
  \caption{Bar chart showing showing K-fold ($K=5$)
    experiments. Although values across folds vary, relative
    performance is consistent.}
  \label{figure:barplot}
\end{figure}

\subsection{Impact of Training / Test Data Split}
Comparing the results of the original HFT paper with the results we
get training over the same split, it becomes clear that models trained
on different splits are not directly comparable.  To further explore
the importance of the chosen split for model selection, we perform
experiments over five randomly selected folds and compare each model
on every fold.

One of the challenges in pursuing empirical work on the Amazon review
dataset is the current absence of a standard train / test split of the
data. Here we evaluate the importance of establishing a standard data
split.

Fig.~\ref{figure:boxplot} shows the results of experiments comparing
the performance of each model over different splits. We perform 5-fold
validation. That is, each model is trained 5 times, each on 80\% of
the data and we report performance on the remaining 20\%. The result
on the test reveal several important points.  First, we note that the
variance over splits can be large, meaning that comparing across
different splits could be misleading when performing model selection.
On the other hand, as shown in Fig~\ref{figure:barplot}, the relative
performance of each model is consistent over the different splits.
This implies that a single random split can be used for model
selection and evaluation as long as this split is held constant over
all evaluated models.

Taken together, Figs. \ref{figure:boxplot} and \ref{figure:barplot}
illustrate the importance of standardizing the dataset splits.
Without standardization, performance measured between different
research groups becomes incomparable and the use of this dataset as a
benchmark is rendered difficult or even impossible.

\begin{figure}
  \includegraphics[width=\columnwidth]{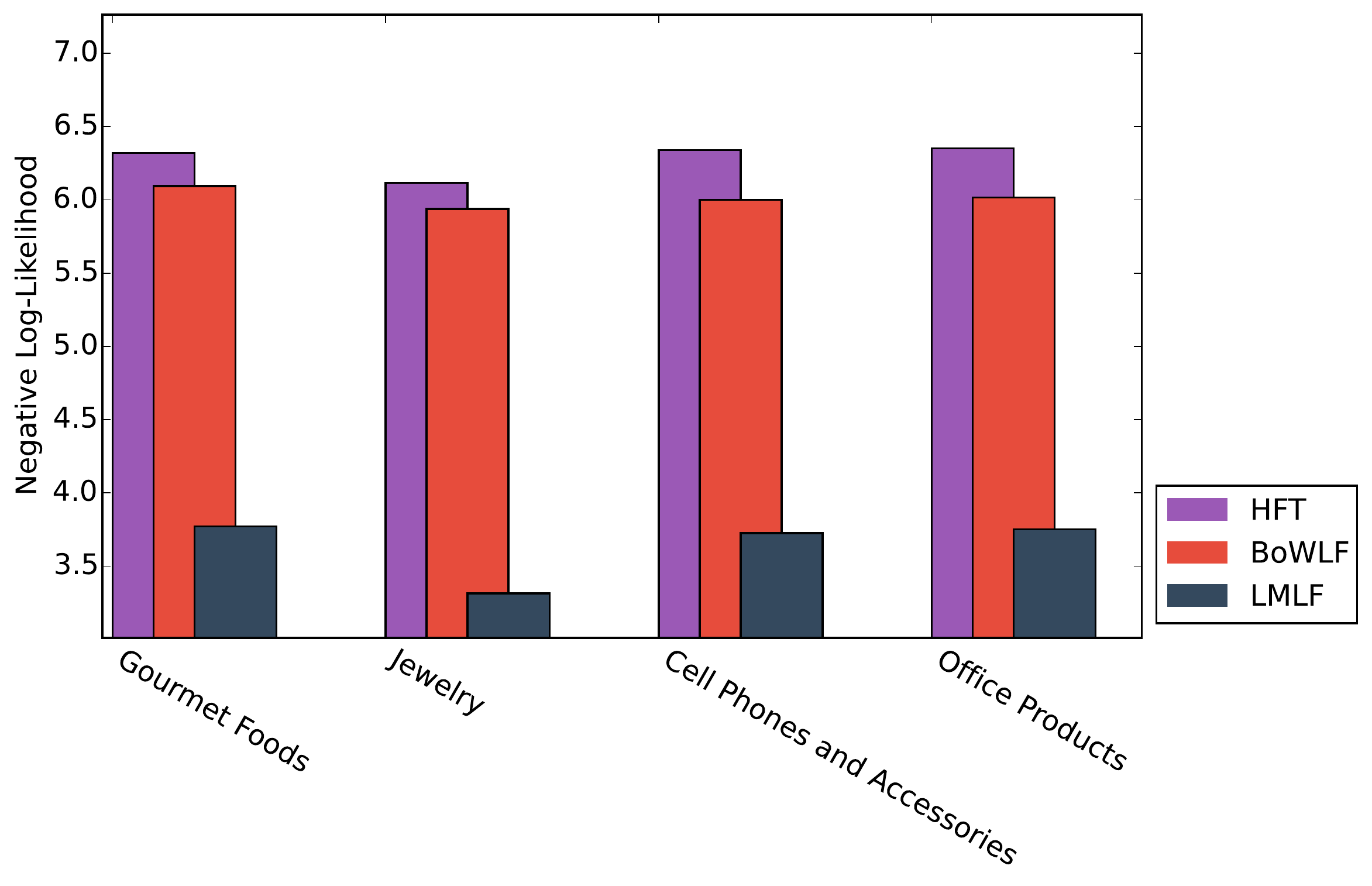}
  \caption{Bar chart showing showing negative log-likelihood (NLL) on
    test data for several datasets. LMLF is superior in NLL but does
    not improve rating prediction over BoWLF.}
  \label{figure:nll}
\end{figure}

\subsection{Effect of Language Model}

One way to analyze models which use text information is to compare
their negative log-likelihood (NLL) scores on the same test dataset.
We find BoWLF has a stronger language model than HFT, which is
reflected in the NLL results, and in this case it appears to
contribute to a better rating prediction. As shown in
Fig.~\ref{figure:nll}, LMLF has a much better language model than both
HFT and BoWLF, but as discussed earlier, LMLF \emph{does not} lead to
better rating predictions than BoWLF.  LMLF appears to be largely
equivalent to HFT in prediction strength, despite having a much better
language model. As discussed above in Sec.~\ref{sec:quant}, this
suggests that the strong nonlinearity in the LSTM helps modeling
reviews, but not necessarily result in the linearly-decodable product
representation, leading to less improvement in rating prediction.

Contrary to LDA-based approaches, the latent dimensions of the product
representations learned by BoWLF do not necessarily have clear
interpretations as topics. However, the neighborhoods learned by BoWLF
are interpretable, where we use the cosine distance between two
product representations, i.e.,
$ 1 - \tfrac{\vgamma_i^\top \vgamma_j}{\|\vgamma_i\|_2
  \|\vgamma_j\|_2}.  $
The BoWLF product space neighbors seem qualitatively superior to the
neighbors given by HFT as seen in Table~\ref{table:neighbors}. Note in
particular the association of ``MTR Simply Tomato Soup'' with other
soups by BoWLF, while HFT neighbors seem much broader, including
crackers, noodles, and gummy bears. This observation is consistent
with the interpretation of the differences in the mathematical form of
HFT and BoWLF (as argued in Sec.~\ref{sec:compare}). The ability of
BoWLF to form peakier distributions over words, given the product
representation, allows the model to be more discriminating and more
closely group similar products. Furthermore, we can see that the
neighbors based on the product representations from the LMLF are
qualitatively worse than those from the BoWLF, which indirectly
confirms that the underlying product-related structure encoded by the
LMLF is more difficult to extract linearly.

While drawing firm conclusions from this small set of neighbors is
obviously ill-advised, the general trend appears to hold in more
extensive testing.  Broadly speaking, this further strengthens the
idea that stronger product representations lead to improvements in
rating prediction.


\begin{table*}[!ht]\

  \small
  \begin{tabular}{l||l|l|l}
    Product                & HFT & BoWLF & LMLF \\
    \hline                                                                              
    \hline                                                                              
    Extra Spearmint & Hong Kong Fu Xiang Yuan Moon Cakes                  & Dubble Bubble Gum                        & Gumballs Special Assorted \\
    Sugarfree Gum                              & French Chew - Vanilla                               & Trident Sugarless White Gum              & Bazooka Bubble Gum        \\
                           & Peck's Anchovette                                   & Gold Mine Nugget Bubble Gum              & Gourmet Spicy Beef Jerky  \\
                           &                                                     &                                          &                           \\
    \hline
    Dark Chocolate                  & Tastykake Kreamies Kakes Ceam .. & Ritter Sport Corn Flakes Chocolate       & Fantis Grape Leaves       \\
    Truffle                              & Miko - Awase Miso Soyabean Paste                    & Chocolate Dobosh Torte                   & Grape Flavoring           \\
                           & Haribo Berries Gummi Candy
                                 & \scriptsize Sugar Free, Milk Chocolate Pecan Turtles & Tutti Fruitti Flavoring   \\
                           &                                                     &                                          &                           \\
    \hline
    MTR Simply & Wellington Cracked Pepper Crackers                  & MTR Mulligatawny Soup                    & Muir Glen Organic Soup    \\
    Tomato Soup                              & Maggi Instant Noodles                               & hai Kitchen Coconut Ginger Soup          & Soy Ginger Saba Noodles   \\
                           & Haribo Gummi Candy                                  & Miko - Awase Miso Soyabean Paste         & Alessi Soup              
  \end{tabular}
  \caption{Nearest neighbors (cosine similarity) based on product representations
    estimated by HFT, BoWLF and LMLF, for Gourmet Foods dataset. Qualitatively,
    the ability to regularize the product
    representations seems to correlate well with the quality of the
    neighbourhoods formed in product representation space.}
  \label{table:neighbors}
\end{table*}

\section{Discussion}

We develop two new models (BoWLF and LMLF) which exploit text reviews
to regularize rating prediction on the Amazon Reviews datasets. BoWLF
achieves state of the art results on 27 of the 28 datasets, while LMLF
outperforms HFT (but not BoWLF) as dataset size increases.
Additionally, we explore the methodology behind the choice of data
split, clearly demonstrating that models trained on different data
subsets cannot be directly compared. Performing K-fold crossvalidation
($K=5$), we confirm that BoWLF achieves superior performance across
dataset splits. The resulting product neighborhoods measured by cosine
similarity between product representations are intuitive, and
correspond with human analysis of the data.  Overall we find that
BoWLF has a $20.29\%$ average improvement over basic matrix
factorization and a $5.64\%$ average improvement over HFT.

We found that the proposed LMLF slightly lagged behind the BoWLF. As
we discuss above, we believe this could be due to the nonlinear nature
of language model based on a recurrent neural network. This
nonlinearity results in the product-related structure underlying
reviews being nonlinearly encoded in the product representation, which
cannot be easily extracted by the linear rating prediction
model. However, this will need to be further investigated in addition
to analyzing the exact effect of language modeling on prediction
performance.


\section*{Acknowledgements}
We would like to thank the developers of Theano
\cite{bergstra+all-Theano-NIPS2011, Bastien-Theano-2012} and Pylearn2
\cite{pylearn2_arxiv_2013}, for developing such a powerful tool for
scientific computing.  We are grateful to Compute Canada and Calcul
Québec for providing us with powerful computational resources.

%
\bibliographystyle{abbrv}
\bibliography{drev_cf,aigaion} 
%
%

\end{document}